\documentclass[twoside,11pt]{article}
\usepackage{jagi}

\usepackage{comment}
\usepackage{fancyvrb}
\usepackage{graphicx}
\usepackage{hyperref}
\usepackage[ragged]{footmisc}
\usepackage{ragged2e}
\usepackage[section]{placeins}
\usepackage{latexsym}
\usepackage{amssymb}

\jagiheading{volume}{issue}{pages}{year}{submitted}{accepted}{DOI}{J G Wolff}

\ShortHeadings{SP as a foundation for the development of human-level AI}{J G Wolff}

\begin{document}

\title{The SP Theory of Intelligence as a Foundation for the Development of a General, Human-Level Thinking Machine}

\author{\name J Gerard Wolff \email jgw@cognitionresearch.org \\
\addr CognitionResearch.org, \\
Menai Bridge, UK}

\editor{Pei Wang}

\maketitle

\begin{abstract}

\noindent This paper summarises how the {\em SP theory of intelligence} and its realisation in the {\em SP computer model} simplifies and integrates concepts across artificial intelligence and related areas, and thus provides a promising foundation for the development of a general, human-level thinking machine, in accordance with the main goal of research in artificial general intelligence.

The key to this simplification and integration is the powerful concept of {\em multiple alignment}, borrowed and adapted from bioinformatics. This concept has the potential to be the ``double helix'' of intelligence, with as much significance for human-level intelligence as has DNA for biological sciences.

Strengths of the SP system include: versatility in the representation of diverse kinds of knowledge; versatility in aspects of intelligence (including: strengths in unsupervised learning; the processing of natural language; pattern recognition at multiple levels of abstraction that is robust in the face of errors in data; several kinds of reasoning (including: one-step `deductive' reasoning; chains of reasoning; abductive reasoning; reasoning with probabilistic networks and trees; reasoning with `rules'; nonmonotonic reasoning and reasoning with default values; Bayesian reasoning with ``explaining away''; and more); planning; problem solving; and more); seamless integration of diverse kinds of knowledge and diverse aspects of intelligence in any combination; and potential for application in several areas (including: helping to solve nine problems with big data; helping to develop human-level intelligence in autonomous robots; serving as a database with intelligence and with versatility in the representation and integration of several forms of knowledge; serving as a vehicle for medical knowledge and as an aid to medical diagnosis; and several more).

\vspace{\baselineskip}


\end{abstract}

\section{Introduction}\label{introduction_section}

% Link to SP software

The {\em SP theory of intelligence} and its realisation in the {\em SP computer model} is the product of an extended programme of research,\footnote{From about 1987 to early 2006, and from late 2012 to the present.} aiming to simplify and integrate observations and concepts in artificial intelligence, mainstream computing, mathematics, and human perception and cognition, with information compression as a unifying theme.

In its quest for simplification and integration, the research accords with Occam's Razor, one of the most widely-accepted principles in science, and it accords with the goal of developing a general, human-level thinking machine, a goal pursued by early researchers in AI and, more recently, by researchers in artificial general intelligence (AGI).

The name ``SP'' derives largely from the fact that information compression---which is central in the workings of the SP system---may be seen as a process of promoting {\em Simplicity} in a body of information by reducing unnecessary complexity in that information whilst retaining as much as possible of its descriptive {\em Power}. The name also relates to the importance, in Occam's Razor, of combining simplicity in a system with its descriptive or explanatory power.

The main aim of this paper is to summarise how the SP system simplifies and integrates concepts across artificial intelligence and related areas, and thus provides a promising foundation for the development of a general, human-level thinking machine, in accordance with the main goal of research in artificial general intelligence.

The key discovery in this research has been that a concept of {\em multiple alignment}, borrowed and adapted from bioinformatics, has great versatility in the representation of diverse kinds of knowledge and great versatility in diverse aspects of intelligence. That two-fold versatility of the multiple alignment framework facilitates the seamless integration of different kinds of knowledge and different aspects of intelligence, an integration that appears to be essential for human-level AI.

I believe it is fair to say that multiple alignment has the potential to be the ``double helix'' of intelligence, with as much significance for human-level intelligence as has DNA for biological sciences.

The next section describes the SP system in outline, while sections that follow aim to demonstrate how the SP system combines relative simplicity with descriptive or explanatory power across areas that are relevant to the development of general, human-level AI.

Since it would not be either feasible or reasonable to repeat what has been published before, this paper is largely a summary of already-published work, with frequent references to relevant sources of information.

\section{Outline of the SP System}\label{outline_of_the_sp_system_section}

The SP system is described most fully in \citet{wolff_2006} and more briefly in \citet{sp_extended_overview}.

Source code for the SP71 computer model accompanies this publication. A very similar and slightly earlier version is described quite fully in \citet[Sections 3.9, 3.10, and 9.2]{wolff_2006}.

\begin{figure}[!hbt]
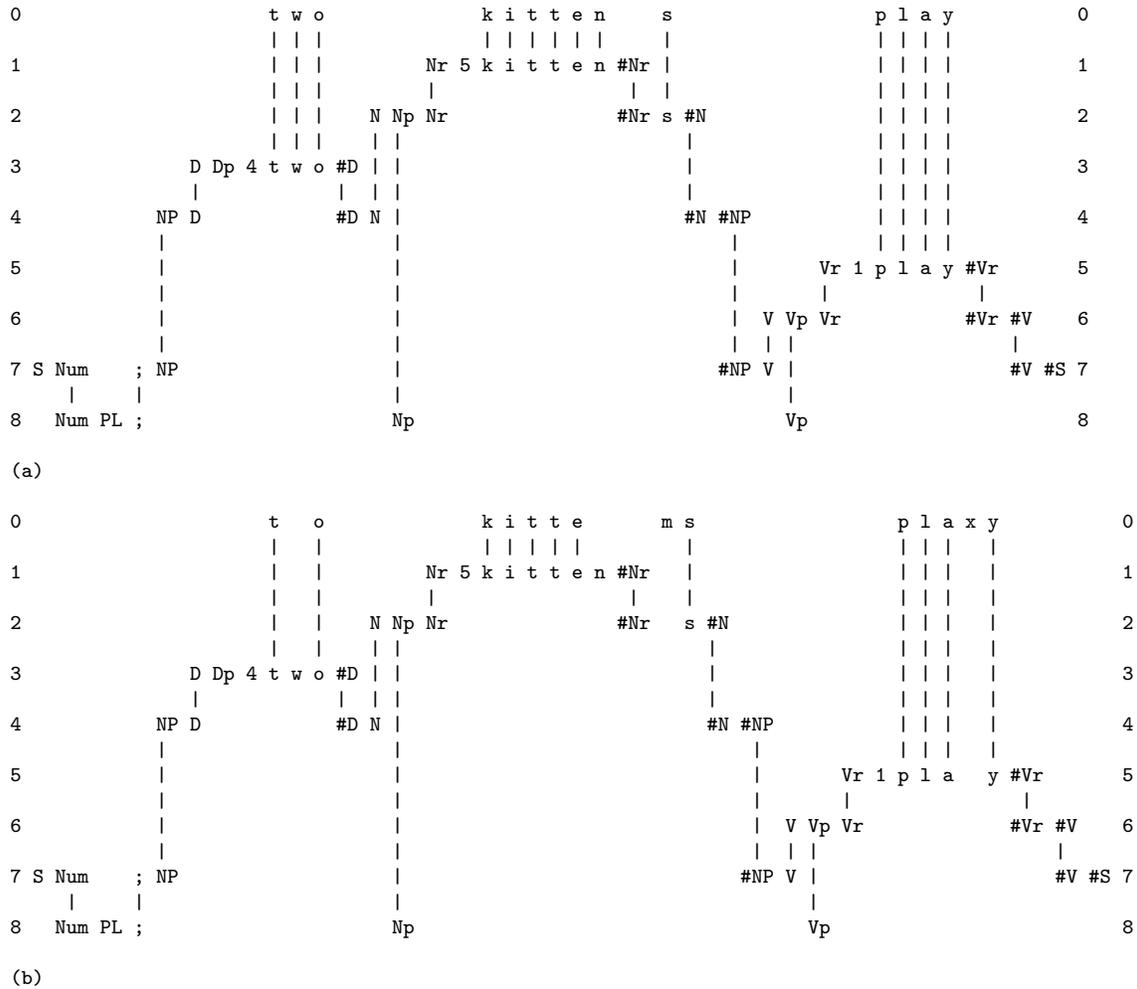

\fontsize{08.00pt}{09.60pt}
\centering
{\bf
\begin{BVerbatim}
0                      t w o              k i t t e n     s                  p l a y           0
                       | | |              | | | | | |     |                  | | | |
1                      | | |         Nr 5 k i t t e n #Nr |                  | | | |           1
                       | | |         |                 |  |                  | | | |
2                      | | |    N Np Nr               #Nr s #N               | | | |           2
                       | | |    | |                         |                | | | |
3               D Dp 4 t w o #D | |                         |                | | | |           3
                |            |  | |                         |                | | | |
4            NP D            #D N |                         #N #NP           | | | |           4
             |                    |                             |            | | | |
5            |                    |                             |       Vr 1 p l a y #Vr       5
             |                    |                             |       |             |
6            |                    |                             |  V Vp Vr           #Vr #V    6
             |                    |                             |  | |                   |
7 S Num    ; NP                   |                            #NP V |                   #V #S 7
     |     |                      |                                  |
8   Num PL ;                      Np                                 Vp                        8

(a)

0                      t   o              k i t t e       m s                  p l a x y           0
                       |   |              | | | | |         |                  | | |   |
1                      |   |         Nr 5 k i t t e n #Nr   |                  | | |   |           1
                       |   |         |                 |    |                  | | |   |
2                      |   |    N Np Nr               #Nr   s #N               | | |   |           2
                       |   |    | |                           |                | | |   |
3               D Dp 4 t w o #D | |                           |                | | |   |           3
                |            |  | |                           |                | | |   |
4            NP D            #D N |                           #N #NP           | | |   |           4
             |                    |                               |            | | |   |
5            |                    |                               |       Vr 1 p l a   y #Vr       5
             |                    |                               |       |               |
6            |                    |                               |  V Vp Vr             #Vr #V    6
             |                    |                               |  | |                     |
7 S Num    ; NP                   |                              #NP V |                     #V #S 7
     |     |                      |                                    |
8   Num PL ;                      Np                                   Vp                          8

(b)
\end{BVerbatim}
}
\caption{a) The best multiple alignment created by the SP computer model with a store of Old patterns like those in rows 1 to 8 (representing grammatical structures, including words) and a New pattern shown in row 0 (representing a sentence to be parsed). Adapted from Figures 1 and 2 in \citet{wolff_sp_intelligent_database}, with permission.}
\label{parsing_kittens_figure}
\end{figure}

The main features of the SP system are these:

\begin{itemize}

    \item All kinds of knowledge are represented with arrays of atomic {\em symbols} in one or two dimensions called {\em patterns}. At present, the SP computer model works only with one-dimensional patterns but it envisaged that the model will be generalised to work with both 1D and 2D patterns.

    \item A ``symbol'' is simply a mark that can be compared with any other symbol to determine whether it is the same or different. Any symbol may have a meaning that derives from its association with other symbols. Otherwise, there is no intrinsic meaning for any symbol, except for some distinctions between different kinds of symbol that serve in the workings of the SP computer model.

    \item The SP system is conceived as a brain-like system that receives {\em New} patterns via its ``senses'' and stores some or all of them in compressed form as {\em Old} patterns.

    \item Processing of information in the SP system is largely a matter of compressing information by searching for patterns, or parts of patterns, that match each other, with a merging or ``unification'' of patterns, or parts of patterns, that are the same. This process---{\em information compression via the matching and unification of patterns}---may be referred to as ``ICMUP'' for short.

    \item More specifically, processing of information in the SP system is largely a process of compressing information by building and processing the previously-mentioned multiple alignments. Two examples of multiple alignments are shown in Figure \ref{parsing_kittens_figure}, below, with some explanation in the text.

    \item Within that general framework, there are two main functions:

    \begin{itemize}

        \item {\em Interpretation of New information via the building of multiple alignments}. Here, the system builds one or more multiple alignments, each of one which provides for the economical encoding one New pattern (sometimes more than one) in terms of one or more Old patterns. Depending on the patterns that are used, this process may be seen to achieve such things as the parsing or understanding or production of natural language, the recognition of patterns, several kinds of reasoning, planning, problem solving, or elements of mathematics.

            Because of the complexity of the search for `good' multiple alignments amongst the many possible `bad' ones, it is necessary to use heuristic methods to ensure that the computational complexity of the process stays within reasonable bounds.

        \item {\em Unsupervised learning of grammars}. In the SP system, unsupervised learning incorporates the process for building multiple alignments. In addition: 1) the system derives newly-created Old patterns from good multiple alignments and adds them, at least temporarily, to its store of Old patterns, and 2) it searches through the store of Old patterns to create one or more collections of Old patterns or {\em grammars} which are relatively effective in the economical encoding of New patterns. As with the building of multiple alignments, the complexity of the process of finding good grammars requires heuristic methods to be applied.

    \end{itemize}

        Although these two functions have here been described separately, they are closely interleaved in practice, in much the same way that human learning is closely interwoven with processes for interpreting what we see, hear, feel, and so on.

    \item Owing to the close connection between information compression and concepts of prediction and probability \citep{li_vitanyi_2014}, the SP system is fundamentally probabilistic.

    \item The SP theory is, at one and the same time, a theory of artificial intelligence, mainstream computing, mathematics, and human learning, perception, and thinking. That said, the main emphasis in work to date has been on aspects of human-level intelligence and, accordingly, the potential of the SP system as a general, human-level thinking machine.

    \item An important part of the theory is {\em SP-neural} (\citet{spneural_2016}, \citet[Chapter 11]{wolff_2006}) which tries to translate abstract concepts like `pattern' and `multiple alignment' in to corresponding concepts in terms of neurons and communications between neurons.

\end{itemize}

Figure \ref{parsing_kittens_figure} (a) shows an example of a multiple alignment that achieves the effect of syntactic analysis or parsing of a sentence, `\texttt{t w o k i t t e n s p l a y}', in terms of grammatical categories including words.

The sentence is shown as a New SP pattern in row 0 and the grammatical categories are shown as Old SP patterns in rows 1 to 8, one pattern per row. This multiple alignment is the best multiple alignment found by the SP computer model when it is supplied with the New pattern and a set of Old patterns representing grammatical categories, including word. Here, `best' means that the multiple alignment is the one out of many that are produced by the SP computer model that achieves the greatest compression of the New pattern in terms of Old patterns.

A point of interest is that the pattern in row 8 of the multiple alignment marks the sentence as having a plural form (`\texttt{PL}'), with an association between a plural subject in the sentence (`\texttt{Np}') with a plural main verb (`\texttt{Vp}'). This way of marking discontinuous dependencies in syntax has an important role in how the SP system may model the subtle and intricate pattern of dependencies in the syntax of English auxiliar verbs (\citet[Sections 8.1, 8.2, and 8.3]{sp_extended_overview}, \citet[Section 5.5]{wolff_2006}).

Figure \ref{parsing_kittens_figure} (b) shows how the SP computer model may produce a plausible parsing despite errors in the New pattern. In this example, there is an error of omission (the absence of `\texttt{w}' in `\texttt{t w o}'), an error of substitution (the word `\texttt{k i t t e n s}' contains `\texttt{m}' instead of `\texttt{n}'), and an error of addition (the word `\texttt{p l a y}' has an added `\texttt{x}').

The robustness of the SP system in the face of errors is not restricted to the processing of natural language. It is a general feature of the SP system that applies in all aspects of the system's workings. The ability of the SP system to tolerate errors in its inputs, within limits, is a by-product of the above-noted probabilistic nature of the system, itself the product of information compression as a central part of how the system works.

The two examples here give only a taste of what can be done with multiple alignment. As noted in the Introduction and outlined below, the multiple alignment concept, as it has been developed in the SP programme of research, is a powerful, versatile vehicle for several different aspects of computing and cognition.

\section{Evaluation of the SP System in Terms of Simplicity and Power}\label{evaluation_section}

This section and the ones that follow summarise evidence that, in accordance with Occam's Razor and the goal of developing a general, human-level thinking machine, the SP theory combines simplicity with explanatory or descriptive power across areas relevant to AGI.

The essential simplicity of the SP system may be seen from two perspectives:

\begin{itemize}

    \item The SP system, as it is expressed in the SP computer model, is largely a process of building multiple alignments and creating grammars, with information compression as a touchstone of success. As such, it is remarkably simple, considering the range of things that it can do, outlined in the sections that follow.

    \item The `exec' file for the SP computer model takes less than 500 kB of storage space.

\end{itemize}

The following sections describe aspects of the SP system that may be seen to be part of its explanatory or descriptive power.

At some stage in the future it may be possible to evaluate a scientific theory like the SP theory with mathematically-precise measures of simplicity and power. Since this is not yet feasible, we have to rely on more traditional assessments in terms of `elegance', `beauty' and, ultimately, the usefulness of the system in the solution of practical problems.

The strengths and potential of the SP system described in the following sections includes features of the system that have been demonstrated already and others that may reasonably be expected with some further development of the system. Relevant evidence is mainly in \citet[Chapters 5 to 10]{wolff_2006}. Other sources of evidence are shown where they apply.

\section{Versatility in the Representation of Knowledge}\label{versatility_in_rk_section}

Although SP patterns are not, in themselves, very expressive, they come to life in the building of multiple alignments. Within that framework, they may serve in the representation of several different kinds of knowledge, listed below.

Here are the main kinds of knowledge that, in evidence to date, may be seen to fall within the scope of the SP system: the syntax of natural language; class hierarchies and class heterarchies (the latter term meaning class hierarchies with cross classification); part-whole hierarchies; discrimination networks and trees; rules for reasoning; structures in mathematics; entity-relationship structures \citep[Sections 3 and 4]{wolff_sp_intelligent_database}; relational knowledge ({\em ibid}., Section 3); patterns in one or two dimensions; images; structures in three dimensions \citep[Section 6.1 and 6,2]{sp_vision}; and procedural knowledge \citep[Section 6.6.1]{sp_benefits_apps}, including parallel procedures \citep[Sections V-G, V-H, and V-I, and Appendix C]{sp_autonomous_robots}.

Since information compression is at the heart of the SP system---in the building of multiple alignments and in unsupervised learning---and since the techniques for the compression of information within the SP system are very general, there is reason to believe that the system may serve in the efficient representation of {\em any} kind of knowledge, not just the examples that have been listed in this section.

In general, the SP system has potential as a {\em universal framework for the representation and processing of diverse kinds of knowledge} (UFK). As such, it has potential to help solve a significant problem with big data: the great variety of formalisms and formats for knowledge and resulting problems in the analysis of big data and for the discovery of associations and structures in big data (Section \ref{benefits_and_applications_section}, \citet[Section III]{sp_big_data}).

\section{Versatility in Intelligence}\label{versatility_in_ai_section}

The building of multiple alignment is not merely a means of giving life to SP patterns in the representation of different kinds of knowledge. It gives expression to several different aspects of intelligence, and it lies at the heart of processes within the SP system for unsupervised learning.

In summary, the following aspects of intelligence fall within the scope of the SP system: unsupervised learning (including the unsupervised learning of segmental structures, classes of structure, and abstract patterns); the parsing and production of natural language with preliminary evidence that the system supports the understanding of natural language and the production of natural language from meanings; pattern recognition that is robust in the face of errors in data (more below); recognition at multiple levels of abstraction; best-match and semantic forms of information retrieval; several kinds of reasoning (more below); planning; and problem solving \citep[Chapters 5 to 10]{wolff_2006}.

The versatility of the process for building multiple alignments may be seen most strikingly in how it gives expression to several different kinds of reasoning. These include: one-step `deductive' reasoning; chains of reasoning; abductive reasoning; reasoning with probabilistic networks and trees; reasoning with `rules'; nonmonotonic reasoning and reasoning with default values; Bayesian reasoning with ``explaining away''; causal reasoning; and reasoning that is not supported by evidence \citep[Chapter 7]{wolff_2006}. It also supports inheritance of attributes in class-inclusion hierarchies and part-whole hierarchies \citep[Section 6.4]{wolff_2006} and there is clear potential for spatial reasoning \citep[Section IV-F.1]{sp_autonomous_robots}, and for what-if reasoning \citep[Section IV-F.2]{sp_autonomous_robots}.

The probabilistic nature of the SP system (Section \ref{outline_of_the_sp_system_section}) means that it can yield plausible results in the face of errors of omission, commission, or substitution in any of its New patterns. This can be seen most clearly in such functions as the parsing of natural language (illustrated in Figure \ref{parsing_kittens_figure} (b)) or the recognition of patterns, but it applies to all aspects of its processing.

There is evidence that the SP system may provide a vehicle not only for the representation of mathematical structures (Section \ref{versatility_in_rk_section}) but for mathematical reasoning and calculations as well. There is a discussion of those possibilities in \citet[Chapter 10]{wolff_2006} and \cite[Section 10]{sp_foundations}.

\section{Seamless Integration of Diverse Kinds of Knowledge and Diverse Aspects of Intelligence}\label{seamless_integration_section}

Multiple alignment, with SP patterns, as a relatively simple vehicle for the representation of diverse kinds of knowledge (Section \ref{versatility_in_rk_section}) and for diverse aspects of intelligence (Section \ref{versatility_in_ai_section}) brings with it a substantial benefit. It means that there is clear potential for the seamless integration of different kinds of knowledge and different aspects of intelligence, in any combination.

It appears that that kind of integration is essential in any system that aspires to achieve human levels of intelligence. Thus it is a major plank in the claim that the SP system provides a promising foundation for the development of a general, human-level thinking machine (Section \ref{introduction_section}).

\section{Potential Benefits and Applications of the SP system}\label{benefits_and_applications_section}

There are several potential benefits and applications of the SP system, some of them---like the nine potential benefits with big data (below)---were quite unexpected. In evaluating the SP system  (Section \ref{evaluation_section}), all the potential benefits and applications are strengths of the system, to be set against its relative simplicity.

Possibilities that have been explored include the following:

\begin{itemize}

    \item {\em Several potential benefits and applications have been outlined in \citet{sp_benefits_apps}}. They include:

    \begin{itemize}

        \item {\em Simplification of computing systems}. There is clear potential for a substantial simplification of computer systems, including software.

        \item {\em Processing of natural language}. In view of the strengths of the SP system in the representation of the syntax of natural language and non-syntactic or semantic forms of knowledge (Section \ref{versatility_in_rk_section}) and in the processing of natural language (Section \ref{versatility_in_ai_section}), there is potential for the system to be applied in the understanding and production of natural language and in translating between natural languages;

        \item {\em Bioinformatics}. Since the multiple alignment concept---a key part of the SP system---has been borrowed from bioinformatics, it is likely that the SP system would prove useful in that area of application;

        \item {\em Managing documentation}. There is potential for the system to be applied in managing the diverse kinds of document in a typical software development project, bearing in mind that each document is likely to come in several parts and sub-parts, and that each part may come in several different versions. Likewise for other projects with similar requirements;

        \item {\em Software engineering}. Since the main elements of a conventional computer program may be mapped on to corresponding structures in multiple alignments \citep[Section 6,7]{sp_benefits_apps}, including the representation of parallel processing \citep[Sections V-G, V-H, and V-I, and Appendix C]{sp_autonomous_robots}, there is potential for the system to be applied in software engineering, both in manual programming of a traditional kind and in automatic or semi-automatic programming via unsupervised learning;

        \item {\em Compression of information}. Since the SP system works entirely via the compression of information, it is likely to prove useful in applications intended for that purpose;

        \item {\em The semantic web}. The strengths of the SP system in the representation of knowledge, in reasoning, in the processing of natural language, in unsupervised learning, and in coping with uncertainty, means that it has potential as an aid to the realisation of the Semantic Web.

        \item {\em Detection of computer viruses}. The strengths of the SP system in pattern recognition (Section \ref{versatility_in_ai_section}) suggest that it is likely to be effective in the detection of computer viruses.

        \item {\em Data fusion}. The strengths of the SP system in information compression suggest that it may serve in ``data fusion'', the merging of streams of data from different sources.

    \end{itemize}

    \item {\em Big data}. The SP theory has the potential to help solve nine problems with big data, as described in \citet{sp_big_data}. In summary:

        \begin{itemize}

            \item {\em Overcoming the problem of variety in big data}. There is potential to reduce all the many formalisms and formats for data to one {\em universal framework for the representation and processing of diverse kinds of knowledge} (UFK).

            \item {\em Learning and discovery}. The SP system has potential for the unsupervised learning or discovery of `natural' structures in data \citep[Section 5.2]{sp_extended_overview}.

            \item {\em Interpretation of data}. As indicated in Section \ref{outline_of_the_sp_system_section}, the SP system has strengths in the interpretation of data in the widest sense that includes parsing of natural language, pattern recognition and more.

            \item {\em Velocity: analysis of streaming data}. The SP system lends itself to an incremental style, assimilating information as it is received, much as people do.

            \item {\em Volume: making big data smaller}. Reducing the size of big data via lossless information compression can yield direct benefits in the storage, management, and transmission of data, and indirect benefits in several other areas outlined in this article.

            \item {\em Additional economies in the transmission of data}. The SP system in conjunction with model-based coding \citep[pp.~139--140]{pierce_1961} has potential for very substantial economies in the transmission of data, first via the unsupervised learning of a {\em grammar} for the kind of data to be transmitted, second, via the distribution of a copy of the grammar to all potential senders and receivers of data, and third, via the {\em encoding} by ``Alice'' of any one body of data such as a TV programme, the {\em transmission} of the encoding to ``Bob'', and his {\em decoding} of the encoded data using the same grammar (\citep{sptrans_2017}, \citet[Section VIII]{sp_big_data}).

            \item {\em Energy, speed, and bulk}. With the SP system, there is potential for big cuts in the use of energy by computers, for greater speed in processing with a given computational resource, and for corresponding cuts in the size and weight of computers.

            \item {\em Veracity: managing errors and uncertainties in data}. The SP system can identify possible errors or uncertainties in data, suggest possible corrections or interpolations, and calculate associated probabilities.

            \item {\em Visualisation}. Knowledge structures created by the system, and inferential processes in the system, are all transparent and open to inspection. They lend themselves to display with static and moving images.

        \end{itemize}

    \item {\em Autonomous robots}. The SP system can help develop human-like versatility and adaptability in autonomous robots (which would be helpful in places like Mars where there is relatively little scope for direct control by people), and the system can help to increase the computational and energy efficiency of robot `brains' and to reduce their size and weight (which are potential problems since each autonomous robot must carry all its intelligence with it and likewise for its energy supplies) \citep{sp_autonomous_robots}.

    \item {\em Intelligent database}. The SP system may serve as a database system with versatility in the representation and integration of several different forms of knowledge and with versatility in several aspects of intelligence which may work together in any combination \citep{wolff_sp_intelligent_database}.

    \item {\em Medical diagnosis}. The SP system may serve as a vehicle for the storage and representation of knowledge about diseases and for the application of that knowledge in medical diagnosis \citep{wolff_medical_diagnosis}. Like other systems of that kind, it would probably be most useful as an assistant to people with medical training rather than as an authority in its own right.

    \item {\em The understanding of natural vision and the development of omputer vision}. The SP system provides an interpretation of several phenomena in natural vision and it provides a framework for the development of computer vision, including scene analysis.

    \item {\em Commonsense reasoning}. A recent paper by Ernest Davis and Gary Marcus \citeyearpar{davis_marcus_2015} argues persuasively that commonsense reasoning---the kind of reasoning we use every day in a wide variety of situations---is surprisingly challenging for artificial intelligence. A paper that I have drafted \citep{csrk_2016} argues, with examples drawn from the Davis and Marcus paper, that the SP system provides a promising foundation for further developments in this area.

\end{itemize}

\section{Other Strengths of the SP System}

The SP system has other strengths that don't fall neatly into the areas outlined in the preceding four sections. These other strengths are described briefly here:

\begin{itemize}

    \item {\em Distinctive features and advantages of the SP theory}. To highlight what is special about the SP theory, its distinctive features are summarised in the first part of \citet{sp_alternatives}. At substantially more length, that paper describes {\em advantages} of the SP theory compared with a range of AI-related alternatives. In particular, Section V of the paper describes 13 problems with deep learning and how, in the SP system, they may be overcome.

    \item {\em Building from primitive foundations}. The SP theory excludes ready-made ideas from mathematics, logic, and computer science, like `equivalence' (`$\Leftrightarrow$'), `implies' (`$\Rightarrow$'), `is-provable' (`$\vdash$'), `{\em variable}', `equals' (`$=$'), `{\em if ...~then}', and so on. This is because the theory is not only a theory of AI, computing, and mathematics, but also a theory of human learning, perception and cognition (Section \ref{outline_of_the_sp_system_section}). Because the primitives in the theory need to be plausible as inborn features of our cognitive machinery, the SP theory has been grounded at a low level that excludes concepts like those mentioned above, which are the products of academic research and which we learn in school or college.

        Of course, the SP theory needs to account for how our minds can accommodate concepts from mathematics, logic, and computer science. How some concepts of those kinds may be modelled in the multiple alignment framework is described in \citet[Section 6.6.1]{sp_benefits_apps}, \citet[Section 10]{sp_foundations}, and \citet[Chapter 10]{wolff_2006}.

    \item {\em Only one kind of processing}. In keeping with the remarks just made, all kinds of processing in the SP system is done via the compression of information by the matching and unification of patterns and, more specifically, via the building and processing of multiple alignments. All kinds of inference that have, so far, been demonstrated with the SP system flow from information compression via the matching and unification of patterns without the need to assume that we are born with higher-level capabilities of the kinds that feature in logic, mathematics, and the like.

    \item {\em The significance of information compression}. In its underlying philosophy, the SP system is consistent with an accumulation of evidence that information compression is significant in both natural cognition and artificial computing:

    \begin{itemize}

        \item {\em Information compression in the workings of brains and nervous systems}. There is a body of research by \citet{attneave_1954}, \citet{barlow_2001_bbs,barlow_1969}, and others, showing the importance of information compression in the workings of brains and nervous systems.

        \item {\em Information compression in learning}. There is a body of evidence that information compression is central in the learning of natural language \citep{wolff_1988}. It appears that the same principles apply to the learning of other kinds of knowledge.

        \item {\em Information compression and generalisation}. Information compression provides a neat answer to the problem of generalisation in machine learning without over-generalisation or under-generalisation \citep[Section 5.3]{sp_extended_overview}. In brief, the product of compressing any given body of information {\em \bf I} may be seen to comprise a grammar and an encoding of {\em \bf I} in terms of the grammar. There is quite good evidence that the grammar provides for generalisation beyond {\em \bf I} without either over-generalisation (underfitting) or under-generaliaation (overfitting).

        \item {\em The DONSVIC principle}. There is evidence that, when machine learning is designed to achieve relatively high levels of information compression, the resulting structures take forms that people regard as natural. This principle---{\em the discovery of natural structures via information compression} (DONSVIC)---provides indirect support for the use of information compression as the foundation for machine learning and other aspects of cognition, as it is in the SP system.

        \item {\em Information compression and prediction and probability}. As noted in Section \ref{outline_of_the_sp_system_section}, there is an intimate connection between information compression and concepts of prediction and probability. This provides support for information compression as the foundation for computing, cognition, and mathematics, since all those things may be seen to be largely about predicting the future from the past. And it may be seen to be the basis for the above-noted robustness of the SP system in the face of errors, something with clear relevance to the natural selection of living things (next).

        \item {\em Information compression in natural selection}. It is likely that, with any organism that stores and uses information as an aid to living, natural selection would favour compression of that information: by reducing the storage space needed for any given body of information, or increasing the effective capacity of any storage space; by reducing the bandwidth needed to transmit any body of information, or increasing what can be sent with a given bandwidth, or increasing speeds of transmission; and perhaps most importantly by allowing the organism to make predictions and calculate probabilities for events in the future \citep[Section 4]{sp_foundations}---and, as mentioned above, to compensate for errors in the system's data, provided they are not too large or numerous.

    \end{itemize}

    \item {\em SP-neural}. Some of the abstract concepts in the SP theory---mainly SP patterns and, less certainly, the building of multiple alignments---map fairly straightforwardly into {\em SP-neural}, a version of the SP theory expressed in terms of neurons, connections between neurons, and communications amongst neurons (\citet{spneural_2016}, \cite[Chapter 11]{wolff_2006}). Although SP-neural is still fairly sketchy and needs to be developed and refined as a working computer model, the fact that one can envisage how abstract concepts in the SP theory may be realised in terms of neurons and their interconnections, lends support to the SP theory as a theory of human learning, perception and thinking as well as being a theory of AI, mainstream computing, and mathematics. The SP theory would be weaker if it were purely abstract without any possibility of seeing how the concepts might be realised in the brain.

        It is not without interest that the concept of a ``pattern'' in the SP theory is quite similar to Donald Hebb's \citeyearpar{hebb_1949} concept of a ``cell assembly'', a concept which itself has evidence in its support and is consistent with a weight of evidence in support of ``grandmother'' cells and localist models of the brain \citep{roy_2013,gross_2002}.

        Although SP-neural is broadly consistent with Hebb's concept of a cell assembly, learning processes in the SP theory overcome a weakness in Hebb's theory of learning, a weakness that Hebb himself recognised. The problem with ``Hebbian'' learning---the gradual strengthening of connections between neurons that frequently fire at the same time---is that it does not explain the commonplace observation that we can and often do learn from a single exposure or experience. The SP theory overcomes this weakness by providing for learning from a single experience and, at the same time, it explains why it takes time to learn complex knowledge and skills---because of the large abstract space that has to be searched to develop solutions that are `good' in terms of information compression.

        Despite its prominent weakness, Hebbian learning has been enthusiastically adopted as the basis for most neural theories of learning and cognition, including deep learning. This is just one of several problems with deep learning, mentioned under the next bullet point.

\end{itemize}

\section{Discussion and Conclusion}\label{conclusion_section}

% Bengio quote.
% Banavar Watson. Kluge v AGI, kluge is more practical? (There is nothing so practical as a good theory). Kluge as a short-term fix for selected problems.
% A kluge may yield practical results quickly, but the SP system may also yield practical results on relatively short timescales.

In this paper, I have tried to show that the SP system, in accordance with Occam's Razor, combines relative simplicity with a relatively large descriptive and explanatory power: in the representation of several kinds of knowledge; in how it accounts for several aspects of intelligence; in the seamless integration of diverse kinds of knowledge and diverse aspects of intelligence---something that appears to be essential in any model that aspires to human-level intelligence; in several potential benefits and applications of the system; and in other aspects that don't fit into the forgoing categories.

I believe that, in its favourable combination of simplicity and power, the SP system clears the path towards the development of a general, human-level thinking machine or artificial general intelligence. Of course there is more to be done but the SP research appears to provide a promising foundation for the development of AGI. As was mentioned in the Introduction, multiple alignment---which is central in the workings of the SP system---has the potential to be the ``double helix'' of intelligence, with as much significance for human-level intelligence as has DNA for biological sciences.

\subsection{In Some Respects, the Human Mind Is a Kluge}

Trying to develop a general, human-level thinking machine is all very well but it seems to be in conflict with quite strong evidence, presented by Gary Marcus \citeyearpar{marcus_2008}, that, in many respects, the human mind is a kluge: ``a clumsy or inelegant---yet surprisingly effective---solution to a problem'' ({\em ibid}., p~2). In view of that evidence, shouldn't the focus of our research be on developing a model of human intelligence, in the spirit of Minsky's ``Society of Mind'' \citep{minsky_1986}, that recognises the messiness of human thinking and takes account of the way that evolution always works with what is at hand so that it may not always arrive at a solution that is theoretically ideal?

The suggestion here is that we need to do both things: we need to recognise that, very likely, the human mind is in many respects a kluge, and we need to continue searching for powerful general mechanisms behind the messiness of the human mind. The latter posture is dictated by the central goal of science---to search for simplicity behind the apparent complexity of the world---but it is also dictated by the failure of AI research over many decades to develop any account of human learning, perception and thinking, either as a kluge or as an elegant general mechanism, that does justice to the versatility and adaptability of human intelligence.

\subsection{We Need More Than One Approach to the Development of Human-Level AI}

As mentioned earlier, I have described in another paper \citep{sp_alternatives} distinctive features of the SP theory compared with AI-related alternatives and, in more detail, advantages of the SP theory compared with a selection of those alternatives.

That analysis may suggest a winner-takes-all approach to the management of research. But, to the contrary, there is already an over-strong tendency for one dominant approach to a problem to squeeze out alternative approaches, witness the relatively enormous publicity and resources currently devoted to research in deep learning, despite the 13 shortcomings of deep learning, with solutions in the SP system, described in \citet[Section V]{sp_alternatives}.

Given the difficulties of achieving human-level AI, we need a multi-pronged strategy. Rather than piling all our eggs into one basket, there should be research on several fronts, including deep learning.

Given the evidence presented in this paper, that multiple alignment may indeed provide the basis for a general, human-level thinking machine, I believe that the SP system is a strong candidate for further research and development.

% \section*{Acknowledgements}

% \appendix

\bibliographystyle{agi}
% \bibliography{latex_references}

\end{document}